\documentclass[letterpaper, 10 pt, conference]{ieeeconf}  

\IEEEoverridecommandlockouts                              

\usepackage{amsmath} 
\usepackage{amssymb}  
\usepackage{graphicx}
\usepackage{multirow}
\usepackage{booktabs}
\usepackage{caption}
\usepackage{subcaption}
\usepackage{bm}
\usepackage{xcolor}
\usepackage{dsfont}
\usepackage{balance} 

\usepackage{longtable}

\usepackage{booktabs}
\usepackage{makecell}
\usepackage{icomma}
\usepackage{tikz}
\usetikzlibrary{tikzmark}

\usepackage{array}
\newcolumntype{L}[1]{>{\raggedright\let\newline\\\arraybackslash\hspace{0pt}}m{#1}}
\newcolumntype{C}[1]{>{\centering\let\newline\\\arraybackslash\hspace{0pt}}m{#1}}
\newcolumntype{R}[1]{>{\raggedleft\let\newline\\\arraybackslash\hspace{0pt}}m{#1}}

\usepackage[hidelinks]{hyperref}
\usepackage[style=ieee,hyperref,natbib=true,backend=bibtex,firstinits,doi=false,%
     mincitenames=1,maxcitenames=2,maxbibnames=99,sorting=none,terseinits=false,hyperref=true]{biblatex}
\bibliography{references.bib}
\renewbibmacro*{bbx:savehash}{}
\defbibheading{bibliography}[\bibname]{\section*{References}}

\newcommand{\irow}[1]{
  \begin{smallmatrix}(#1)\end{smallmatrix}%
}

\title{\LARGE \bf
Efficient Representations of Object Geometry for\\ Reinforcement Learning of Interactive Grasping Policies
}

\author{Malte Mosbach* and Sven Behnke
\thanks{$^{*}$All authors are with the Autonomous Intelligent Systems group of University of Bonn, Germany; {\tt mosbach@ais.uni-bonn.de}}%
}

\begin{document}

\maketitle
\thispagestyle{empty}
\pagestyle{empty}

\begin{abstract}

Grasping objects of different shapes and sizes---a foundational, effortless skill for humans---remains a challenging task in robotics. Although model-based approaches can predict stable grasp configurations for known object models, they struggle to generalize to novel objects and often operate in a non-interactive open-loop manner. In this work, we present a reinforcement learning framework that learns the interactive grasping of various geometrically distinct real-world objects by continuously controlling an anthropomorphic robotic hand. We explore several explicit representations of object geometry as input to the policy. Moreover, we propose to inform the policy implicitly through signed distances and show that this is naturally suited to guide the search through a shaped reward component. Finally, we demonstrate that the proposed framework is able to learn even in more challenging conditions, such as targeted grasping from a cluttered bin. Necessary pre-grasping behaviors such as object reorientation and utilization of environmental constraints emerge in this case. Videos of learned interactive policies are available at \url{https://maltemosbach.github.io/geometry_aware_grasping_policies}.

\end{abstract}

\section{Introduction}
Grasping is a fundamental capability and the basis for almost all complex manipulation skills, such as pick-and-place or tool use~\cite{Kumar2019}. Consequently, object grasping is of great practical relevance and remains an active area of research. Classical approaches have tackled grasping via analytical, model-based planning and control under the assumption of accurate state estimates and a given dynamics model~\cite{Kumar2014, Mordatch2012}. Recently, reinforcement learning (RL) has grown in popularity for grasping~\cite{Joshi2020, Mohammed2020} and object manipulation~\cite{Akkaya2019, Rajeswaran2017}, due to its ability to autonomously discover useful behaviors.
The rise of massively parallelized physics simulation~\cite{Makoviychuk2021} and the resulting speedup in RL training has led to a leap in the problem complexity for which proficient controllers can be learned, as evidenced by dexterous in-hand manipulation~\cite{Chen2022} as well as agile and robust quadruped locomotion~\cite{Rudin2022, Margolis2022}. Policies are trained in thousands of environments in parallel from low-dimensional environment states, such as joint or object positions. Surprisingly, skillful in-hand manipulation of geometrically distinct objects can be achieved solely through data-driven learning without access to shape or tactile information~\cite{Chen2022}. 

However, in-hand manipulation is so far only concerned with objects of relatively constant size. Objects encountered in grasping and bin-picking scenarios might vary considerably in their extent. It is therefore crucial for a policy to generalize across variations in object shape and size. The effect of object representation on a policy's robustness to variations in shape and size has not been thoroughly investigated, though. Moreover, in-hand manipulation provides dense rewards, as the hand is constantly manipulating and moving the object, achieving states that are closer to or further away from a desired pose. In contrast, object grasping allows the agent to move the hand freely in space. Hence, the majority of random exploration does not change the state of an object and therefore does not provide useful feedback. 

\begin{figure}[t]
      \centering
     
     \begin{subfigure}[b]{0.325\columnwidth}
         \centering
         \includegraphics[width=\textwidth]{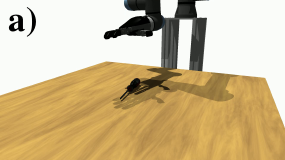}
     \end{subfigure}
     \hfill
     \begin{subfigure}[b]{0.325\columnwidth}
         \centering
         \includegraphics[width=\textwidth]{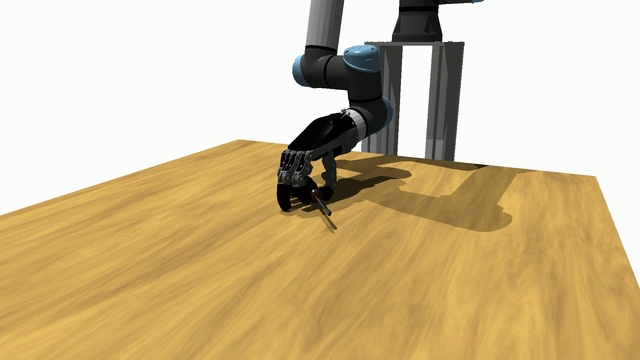}
     \end{subfigure}
     \hfill
     \begin{subfigure}[b]{0.325\columnwidth}
         \centering
         \includegraphics[width=\textwidth]{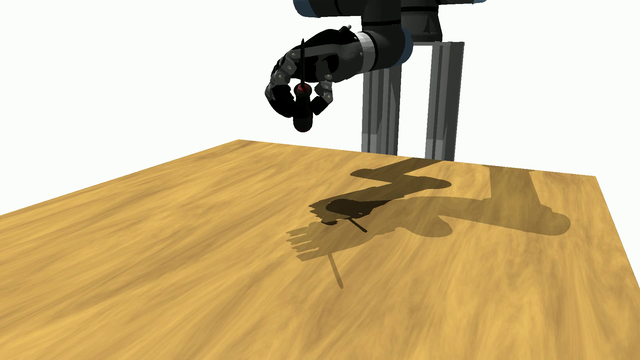}
     \end{subfigure}
     
     \vspace{0.75cm}
     
     \begin{subfigure}[b]{0.325\columnwidth}
         \centering
         \includegraphics[width=\textwidth]{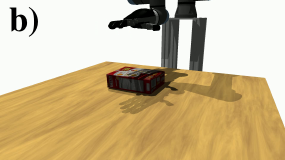}
     \end{subfigure}
     \hfill
     \begin{subfigure}[b]{0.325\columnwidth}
         \centering
         \includegraphics[width=\textwidth]{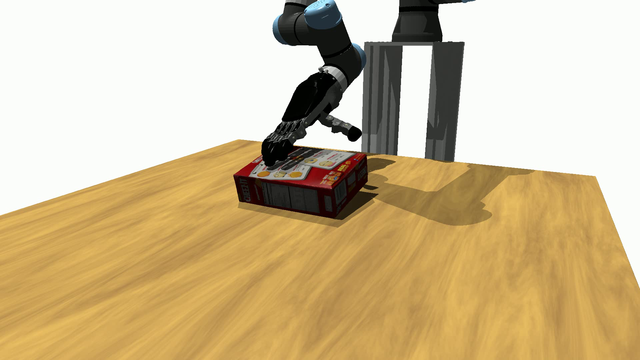}
     \end{subfigure}
     \hfill
     \begin{subfigure}[b]{0.325\columnwidth}
         \centering
         \includegraphics[width=\textwidth]{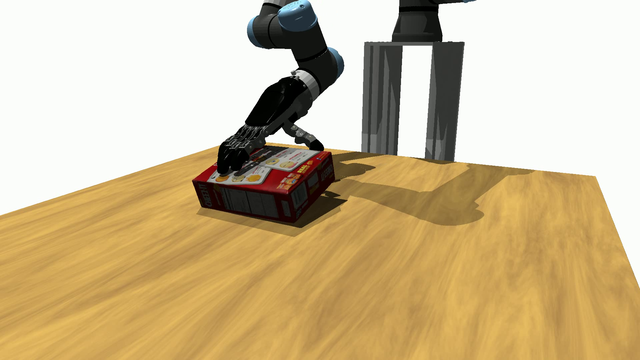}
     \end{subfigure}
     
     \vspace{0.1cm}
     
     \begin{subfigure}[b]{0.325\columnwidth}
         \centering
         \includegraphics[width=\textwidth]{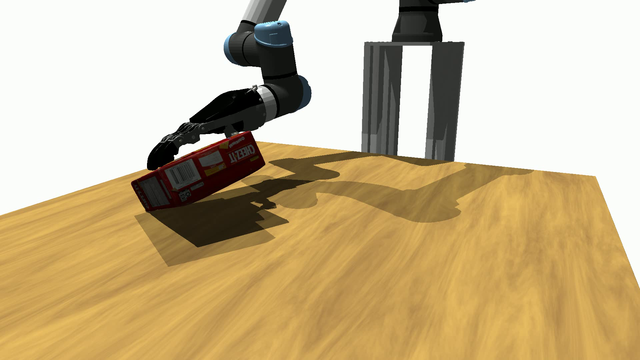}
     \end{subfigure}
     \hfill
     \begin{subfigure}[b]{0.325\columnwidth}
         \centering
         \includegraphics[width=\textwidth]{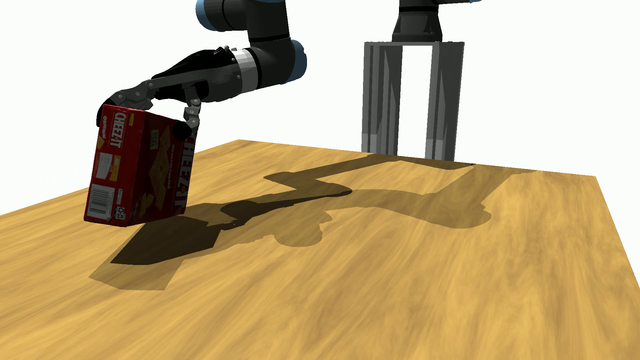}
     \end{subfigure}
     \hfill
     \begin{subfigure}[b]{0.325\columnwidth}
         \centering
         \includegraphics[width=\textwidth]{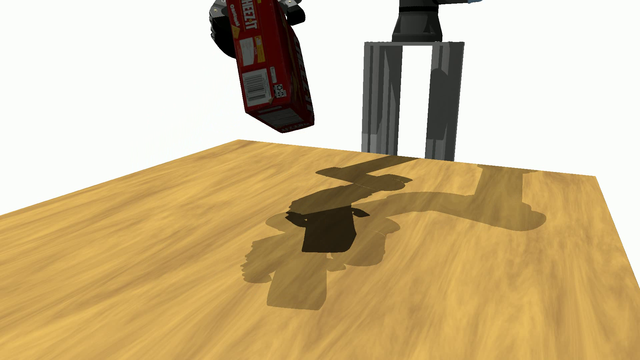}
     \end{subfigure}
     
     \vspace{0.75cm}
     
     \begin{subfigure}[b]{0.325\columnwidth}
         \centering
         \includegraphics[width=\textwidth]{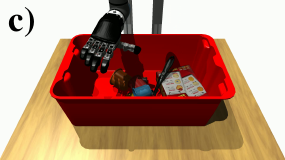}
     \end{subfigure}
     \hfill
     \begin{subfigure}[b]{0.325\columnwidth}
         \centering
         \includegraphics[width=\textwidth]{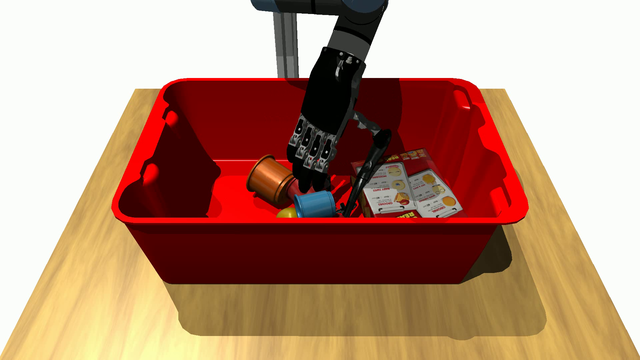}
     \end{subfigure}
     \hfill
     \begin{subfigure}[b]{0.325\columnwidth}
         \centering
         \includegraphics[width=\textwidth]{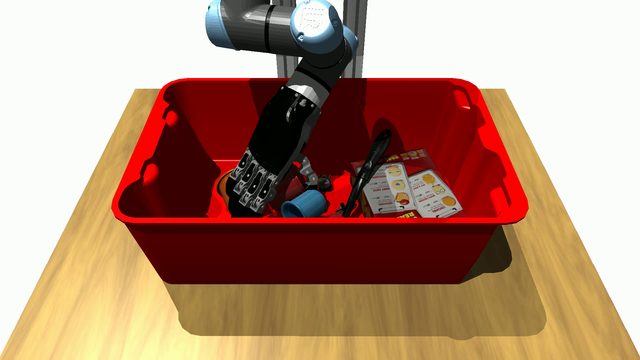}
     \end{subfigure}
     
     \vspace{0.1cm}
     
     \begin{subfigure}[b]{0.325\columnwidth}
         \centering
         \includegraphics[width=\textwidth]{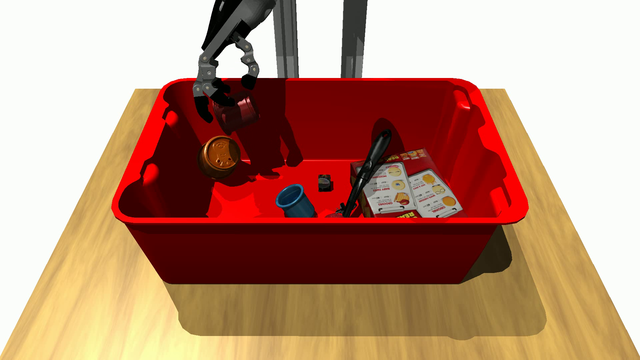}
     \end{subfigure}
     \hfill
     \begin{subfigure}[b]{0.325\columnwidth}
         \centering
         \includegraphics[width=\textwidth]{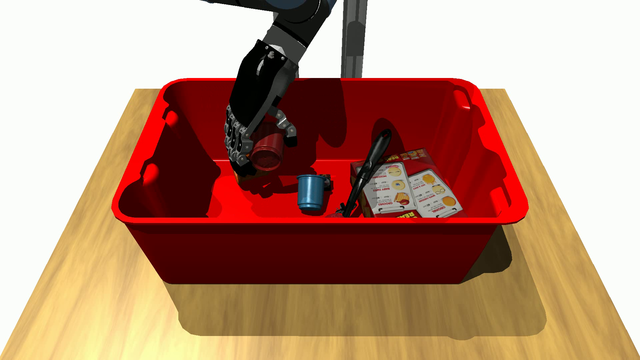}
     \end{subfigure}
     \hfill
     \begin{subfigure}[b]{0.325\columnwidth}
         \centering
         \includegraphics[width=\textwidth]{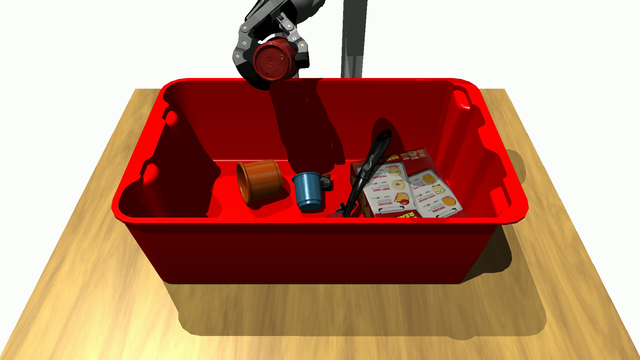}
     \end{subfigure}

      \caption{A single policy learns to grasp objects of vastly different shapes, weights and sizes: (a) Screwdriver that can be readily picked up; (b) Larger cracker box, impossible to grasp from the initial configuration. The policy, nevertheless, learns to lift the object through dexterous pre-grasp manipulation that brings the box into an upright position; (c) Red cup is retrieved from a cluttered container, even as it interacts with other objects and the bin.}
      \label{grasping_rollouts}
\end{figure}

In this work, we investigate whether the representation of objects via their center-of-mass is sufficient for our problem, or whether representations that explicitly or implicitly encode object shape can lead to more capable policies. While visual perception, e.g., via RGB-D images, provides detailed information about object geometry, we attempt to find more concise representations. This is motivated because i) learning directly from high-dimensional visual input is more challenging, leading to greatly increased sample complexity, and ii) rendering visual observations in many parallelized environments quickly fills GPU memory, reducing the number of simulations that can be run in parallel and thus undermining the very benefit of massively parallelized training. To avoid these issues, we investigate popular explicit representations of 3D geometries for their performance in RL. Furthermore, we utilize the signed distance function (SDF) to implicitly inform an agent about an object's shape. 
We address the sparse reward problem by proposing a novel geometry-aware shaped reward that can achieve the desired reduction of the search space, without biasing the solutions that are found. 

We observe that with these representations, existing RL algorithms are able to learn resilient grasping controllers that automatically exhibit pre-grasp manipulation and regrasping behaviors. This allows applying the proposed framework to the more challenging task of targeted picking from a cluttered bin. Notably, even without representation of the context, i.e., other objects and the bin, we are able to learn proficient grasping policies for this scenario.
In summary, we make the following contributions:

\begin{itemize}
 \item  A study of explicit shape representations for object manipulation. We investigate different representations of object shape with respect to their RL performance. 
 \item  Implicit representation via SDFs is applied to RL of grasping, which has, to the best of our knowledge, not been done in any prior work. From this representation, we infer a geometry-aware reward component encouraging the agent to make contact with an object.
 \item Picking from cluttered bins is learned without information about the context. We demonstrate that proficient policies can be trained even for this unstructured setting.
\end{itemize}

\section{Related Work}

The field of robotic grasping has been an active area of research for many decades. Despite these sustained efforts, the problem remains largely unsolved~\cite{Kleeberger2020}. Approaches to robotic grasping are typically categorized into analytical and data-driven methods. The former, often also referred to as classical approaches to grasping, is concerned with finding suitable grasp poses for an object, where a kinematically feasible path can then be planned to execute the picking action. The quality of possible grasp poses is assessed through mechanical and geometric conditions, such as force closure or form closure~\cite{Zhang2022}. Data-driven methods have been gaining prominence for robotic grasping due to significant increases in performance as more data and computation becomes available. These subsume approaches for learning-based grasp-synthesis, where only the final grasp pose is learned~\cite{Tremblay2018, Dong2019}, as well as dexterous manipulation, which typically involves continuous control of the end-effector. Most recent works use reinforcement learning to generate such interactive policies. Rajeswaran et al.~\cite{Rajeswaran2017} combine model-free RL with human demonstrations to learn dexterous manipulation tasks. DQN is used by Deng et al.~\cite{Deng2019} to learn a policy that picks objects from clutter using a combination of a suction cup and two-fingered gripper. Quillen et al.~\cite{Quillen2018} conduct an empirical evaluation of off-policy model-free RL for vision-based grasping with a parallel gripper.

While the representation of object geometry is a prevalent topic in computer vision, few works address it explicitly in the context of RL. Huang et al.~\cite{Huang2021} study the effect of geometry-awareness for RL of in-hand reorientation and show that a generalist, multi-object policy can be learned that outperforms the single-object baselines. Kumar et al.~\cite{Kumar2019} represent different objects through their bounding boxes. They show that tactile perception is able to compensate for information lost by this approximation as well as measurement noise. In a recent study, Wu et al.~\cite{Wu2022} use both, an imitation learning and a geometric representation learning objective, to train grasping and manipulation policies from point clouds. However, their framework relies on category-specific grasping demonstrations.

Recent work by Cai et al.~\cite{Cai2022} utilizes the truncated SDF (TSDF) for contact point detection. They demonstrate visual grasping from a cluttered bin with a two-fingered gripper. Breyer et al.~\cite{Breyer2020} propose a volumetric grasping network that operates directly on a TSDF representation of a cluttered scene and outputs grasp poses and qualities.  To the best of our knowledge, no prior work has explored the signed distance function for its potential in RL for object manipulation.

In this work, we perform a comparative study of explicit object representations for RL-based grasping. Further, we propose to implicitly represent object geometry through the SDF, with leads to succinct states amenable to large-scale optimization and which naturally lend themselves to rewarding behaviors useful for object grasping.  
\begin{figure}[t]
      \centering
      \begin{subfigure}[b]{0.45\columnwidth}
         \centering
         \includegraphics[width=\textwidth,trim={0 4cm 0 1cm},clip]{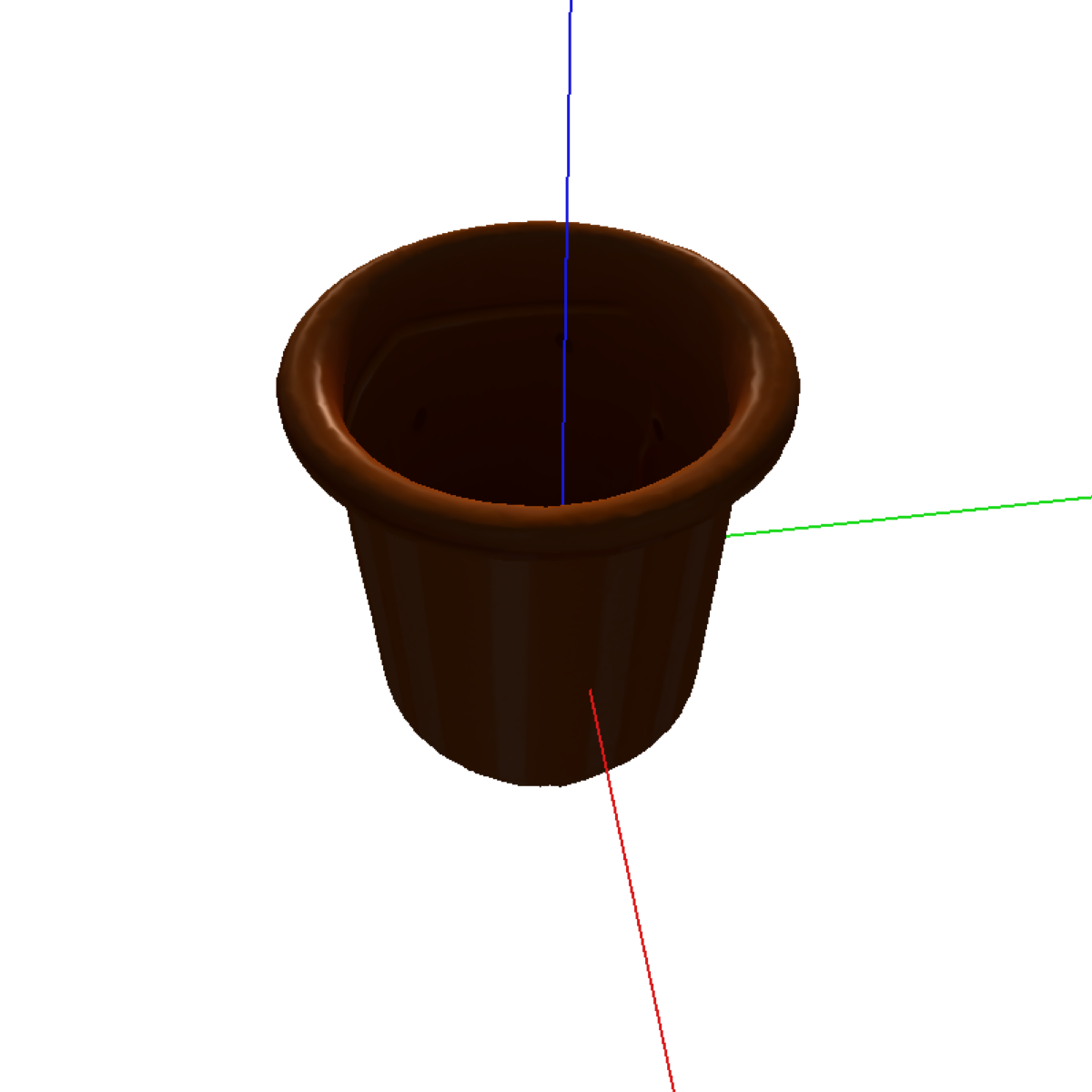}
         \caption{Center of mass (COM)}
         \label{object_representations:com}
     \end{subfigure}
     \hfill
     \begin{subfigure}[b]{0.45\columnwidth}
         \centering
         \includegraphics[width=\textwidth,trim={0 4cm 0 1cm},clip]{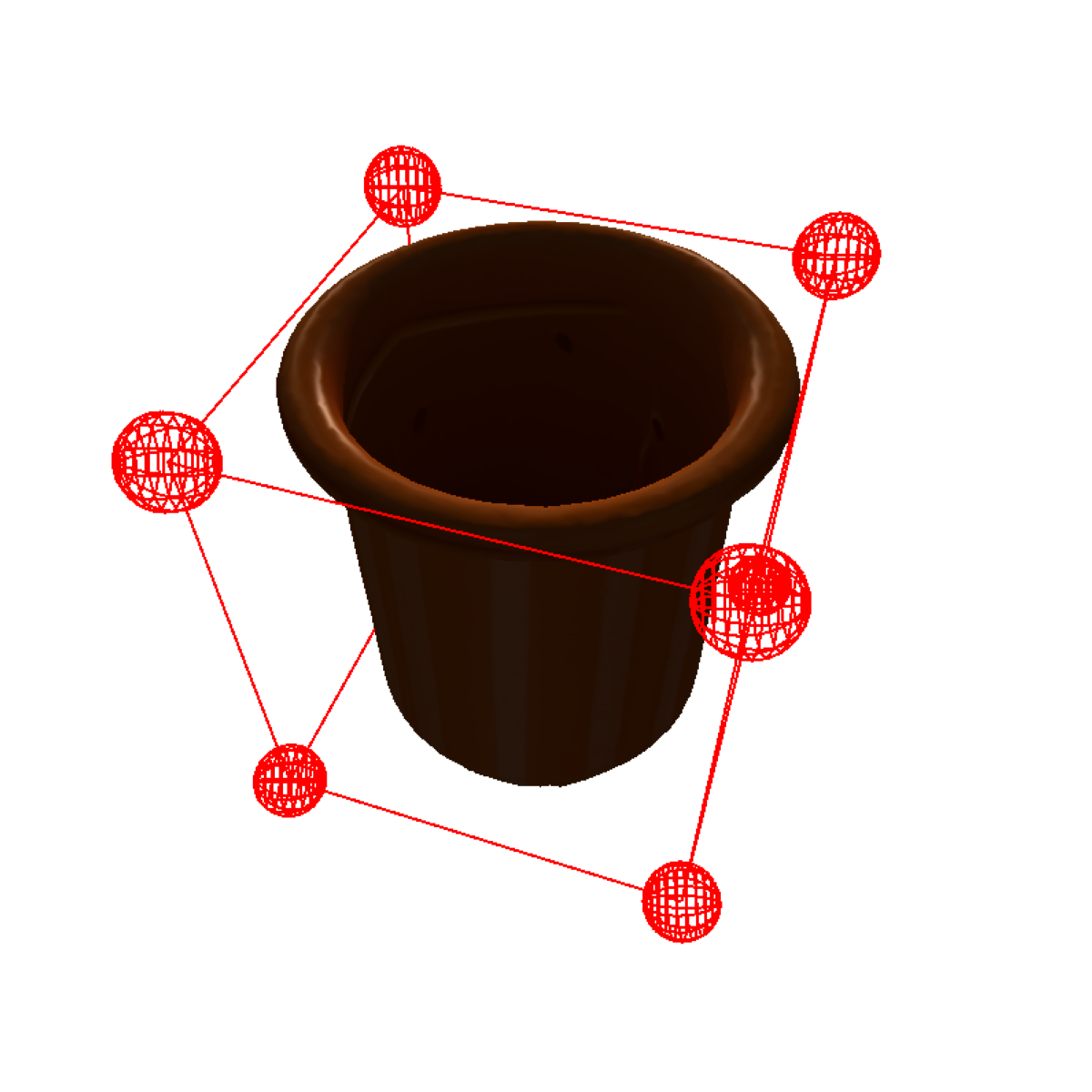}
         \caption{Bounding box (BBox)}
         \label{object_representations:bbox}
     \end{subfigure}
     
     \begin{subfigure}[b]{0.45\columnwidth}
         \centering
         \includegraphics[width=\textwidth,trim={0 8cm 0 2cm},clip]{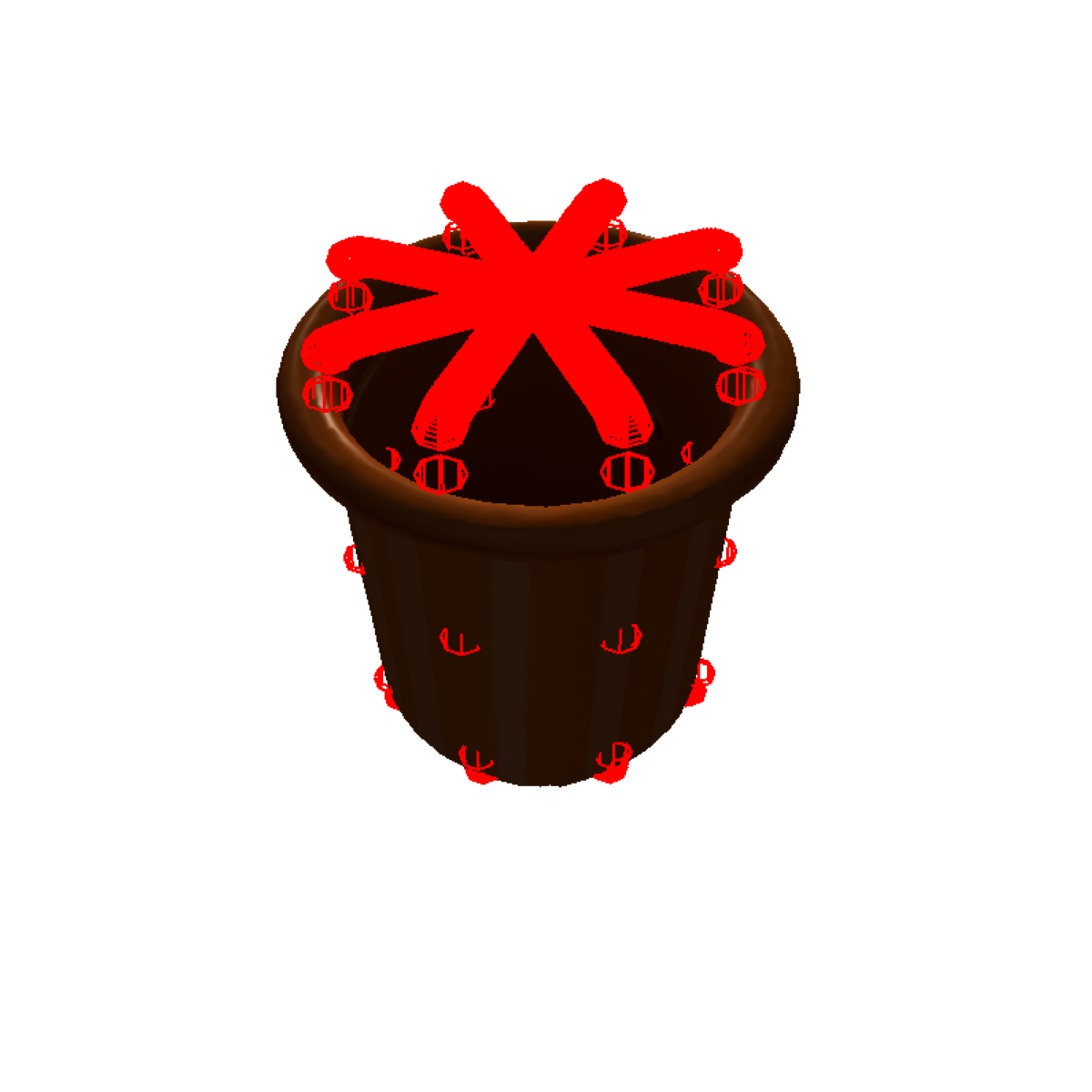}
         \caption{Superquadric (SQ)}
         \label{object_representations:sq}
     \end{subfigure}
     \hfill
     \begin{subfigure}[b]{0.45\columnwidth}
         \centering
         \includegraphics[width=\textwidth,trim={0 8cm 0 2cm},clip]{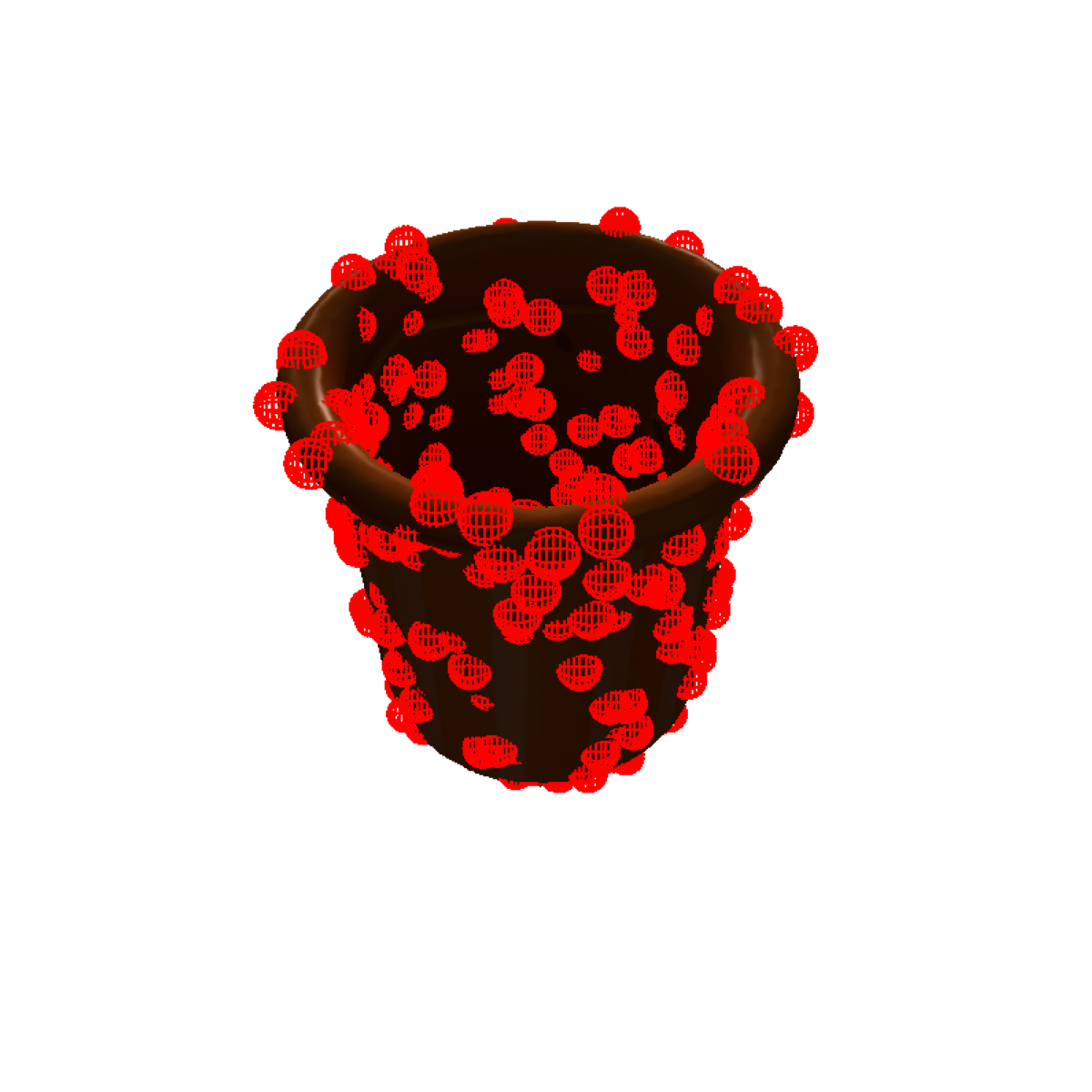}
         \caption{Point cloud (PC)}
         \label{object_representations:pc}
     \end{subfigure}
      
      \caption{We examine four explicit representations of object geometry that differ in dimensionality and expressiveness.}
      \label{object_representations}
\end{figure}

\section{Method}
We formalize the task of grasping an object as a Markov decision process $\mathcal{M} = (\mathcal{S}, \mathcal{A}, R, P, \gamma)$, where $\mathcal{S}$ and $\mathcal{A}$ are the state and action space, respectively. The transition function $P: \mathcal{S} \times \mathcal{A} \times \mathcal{S} \rightarrow \mathbb{R}_{+}$ represents the probability of transitioning to the next state $\bm{s}_{t+1} \in \mathcal{S}$ when taking action $\bm{a}_{t} \in \mathcal{A}$ in the current state $\bm{s}_{t} \in \mathcal{S}$. $R: \mathcal{S} \times \mathcal{A} \rightarrow \mathbb{R}$ is a reward function and $0 \leq \gamma < 1$ is a discount factor.

Our goal is to learn a policy $\pi_{\theta}$ that can grasp a large number $N$ of geometrically diverse objects. This induces a multi-task problem, the goal of which can be formalized as optimizing the expected sum of discounted rewards over all $N$ objects:

\begin{equation}
  \mathbb{E}_{\pi_{\theta}} \left[ \sum_{i=1}^{N}\sum_{t=0}^{T-1} \gamma^{t} R(s_t, \pi_{\theta}(s_t))\right].
\end{equation}

As we will show, naive optimization for this objective without incorporating object-specific information can be surprisingly successful if sufficient amounts of experience can be generated. There is, however, one major drawback to this approach. As the policy cannot distinguish different objects from each other, a common, universally effective strategy must be learned~\cite{Huang2021}. This puts an upper limit to the dexterity that can emerge during optimization. Instead, we aim to learn a policy that adapts to the different objects it encounters and is able to employ specialized, geometry-aware strategies to grasp them. Thus, we investigate representations of object shape for their performance in RL of grasping policies. 

\begin{figure}[t]
      \centering
      \begin{subfigure}[b]{0.475\columnwidth}
         \centering
         \includegraphics[width=\textwidth]{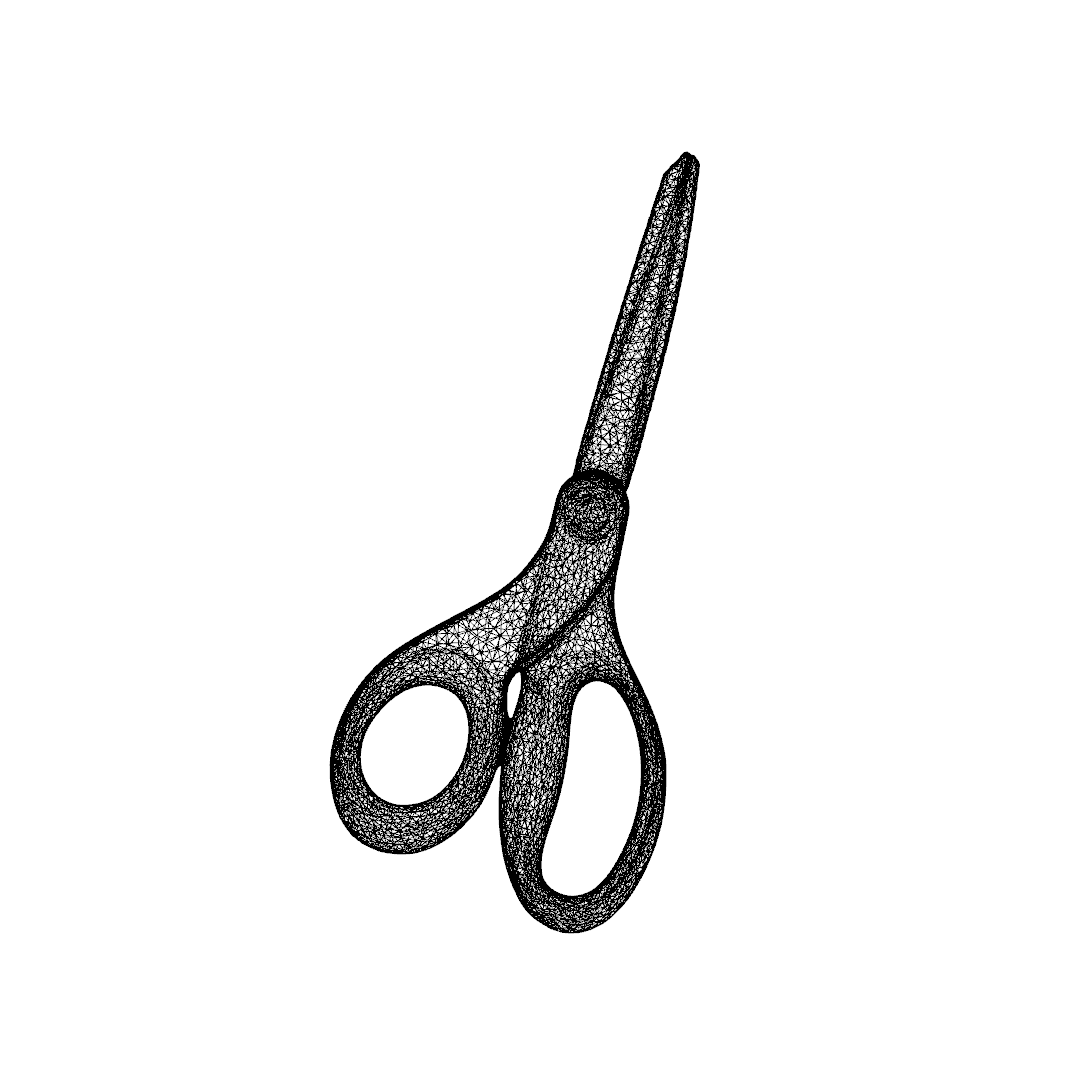}
         \caption{Detailed triangular mesh of an object}
         \label{sdf:mesh}
     \end{subfigure}
     \hfill
     \begin{subfigure}[b]{0.475\columnwidth}
         \centering
         \includegraphics[width=\textwidth]{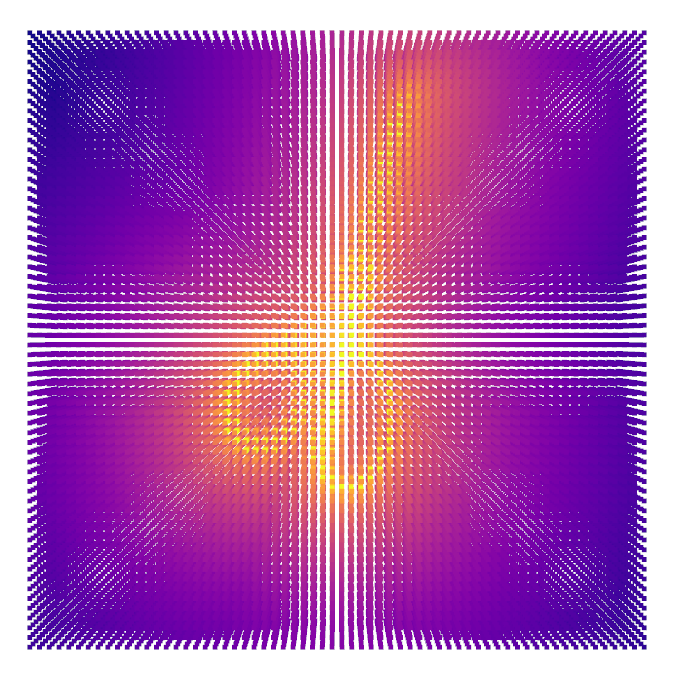}
         \caption{SDF representation of the object saved as a voxel grid}
         \label{sdf:sdf_grid}
     \end{subfigure}
      
      \caption{We precompute voxelized SDFs for object meshes to enable fast lookups of distances during training.}
      \label{sdf}
\end{figure}

\subsection{Explicit Representation of Object Geometry}
The explicit representations of an object's geometry we explore in this work are depicted in Figure~\ref{object_representations}.

\subsubsection{Center of Mass} The naive representation of objects via their center of mass position or 6D pose is prevalent in reinforcement learning of single-object manipulation policies~\cite{Plappert2018, Huang2021}. We employ this representation as a geometry-agnostic baseline. 

\subsubsection{Bounding Box} 3D bounding boxes represent the cuboid hull in which an object lies. We extract the oriented bounding box along the three principal axes of a given mesh. While prior work \cite{Kumar2019} has studied the effect of bounding box representations for grasping, this has only been investigated for contextual RL. To provide a succinct observation to the policy, we represent the bounding box via its central pose and extent.

\subsubsection{Superquadric} Superquadrics go beyond the standard 3D cuboid representation. They are an 11-dimensional parametric family of shapes able to represent cubes, cylinders, spheres, etc.~\cite{Barr1981}. Due to their capacity to represent a variety of shapes with very few parameters, superquadrics have been used as object representations for manipulation~\cite{Silva2016, Vezzani2017} and shape parsing~\cite{Paschalidou2019}. We are, however, not aware of prior work using superquadrics in the context of RL. The section \ref{sec:superquadric_recovery} outlines how we recover superquadric parameters for an object.

\subsubsection{Point Cloud} Point clouds are a natural representation of 3D objects as they can be created by laser scanners or RGB-D cameras. However, they are irregular in the sense that there is no natural ordering of the points. To respect the permutation invariance of the input, we preprocess a point cloud with a PointNet-like~\cite{Qi2017} architecture before feeding the embedding into the MLP policy. We perform experiments for point clouds of different sizes, sampling 32, 128, and 512 points on the mesh surface, respectively.

\begin{figure}[t]
      \centering
      \includegraphics[width=7.813cm]{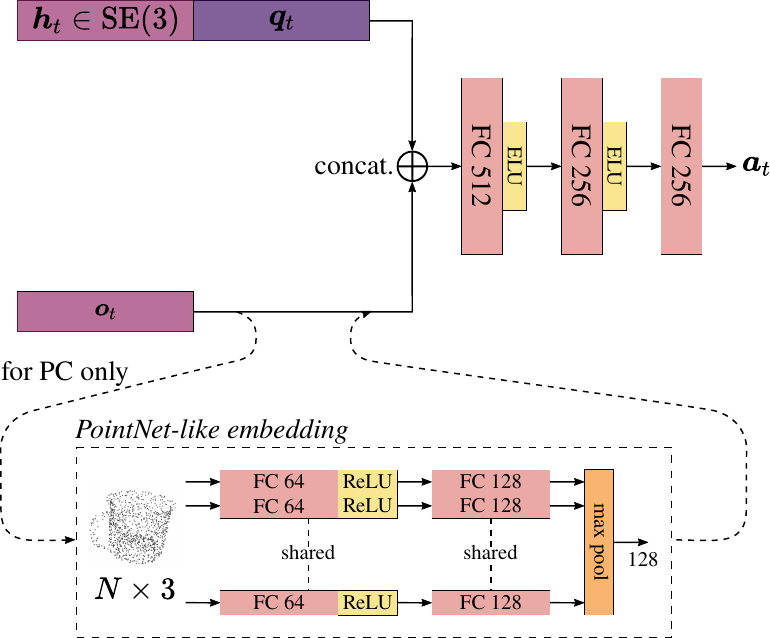}
      \caption{Policy architecture. $\bm{h}_t$ is the hand pose, $\bm{q}_t$ are the joint positions of the hand, and $\bm{o}_t$ is the object representation at time step $t$. When the object is represented as a point cloud (PC), we preprocess the input via a PointNet-like architecture~\cite{Qi2017} to generate a permutation invariant embedding first. The three-layer fully connected neural network outputs actions $\bm{a}_t$ that are changes to the 6\,DoF hand pose and to the five controllable DoF of an anthropomorphic hand.}
      \label{policy_architecture}
\end{figure}

\subsection{Superquadric Recovery}
\label{sec:superquadric_recovery}
\begin{figure}[t]
      \centering
      \begin{subfigure}[b]{0.475\columnwidth}
         \centering
         \includegraphics[width=\textwidth]{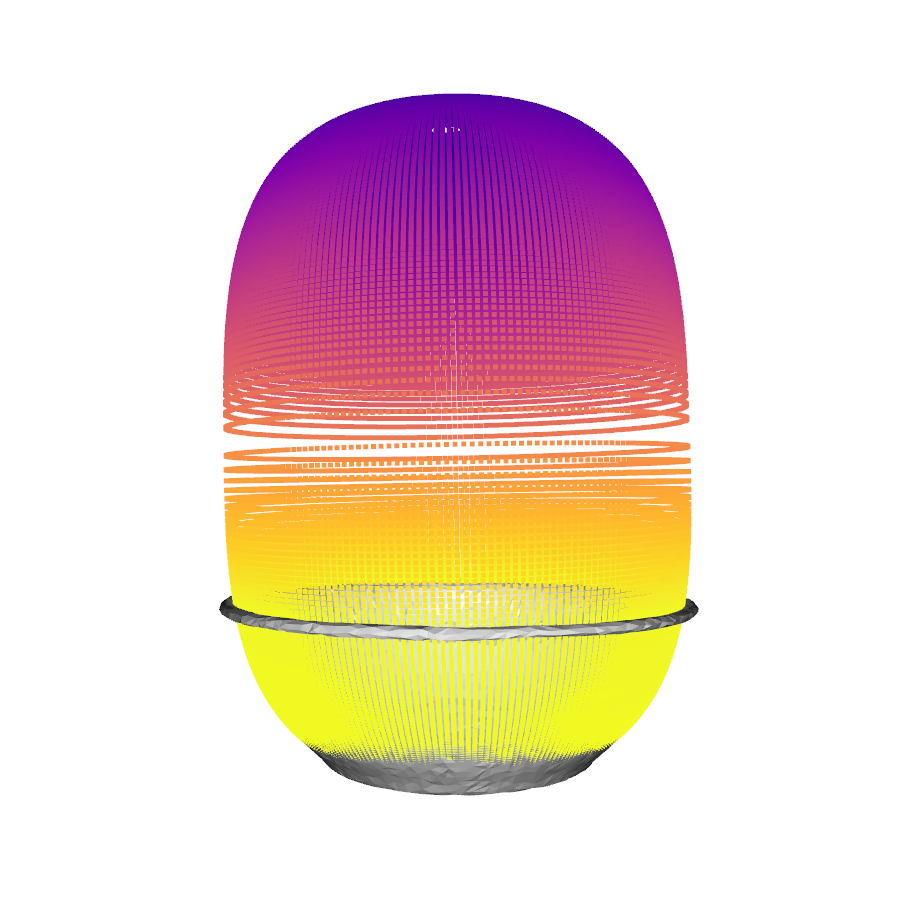}
         \caption{Solution found by the original EMS algorithm}
         \label{sq_samples_fitting:original}
     \end{subfigure}
     \hfill
     \begin{subfigure}[b]{0.475\columnwidth}
         \centering
         \includegraphics[width=\textwidth]{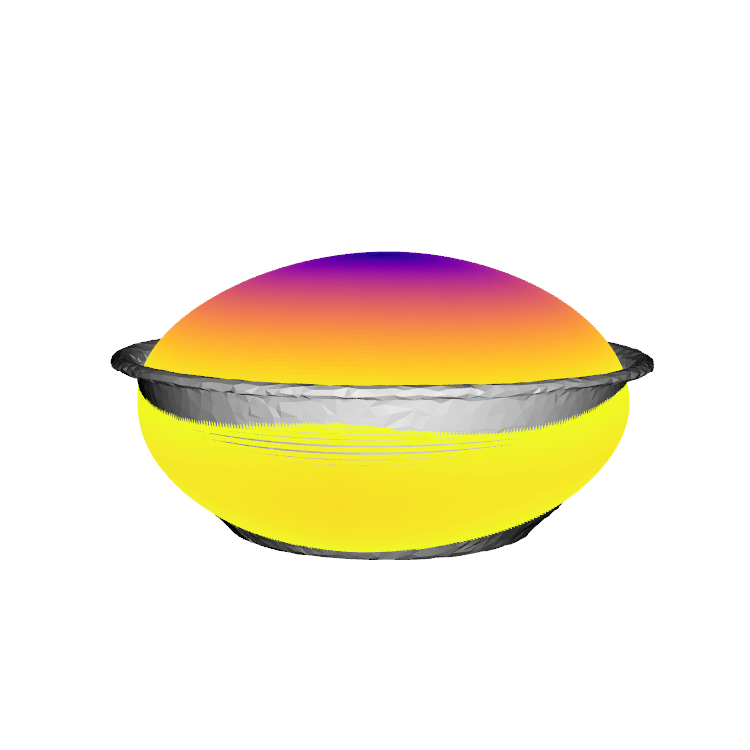}
         \caption{Solution found using our regularization}
         \label{sq_samples_fitting:ours}
     \end{subfigure}
      
      \caption{We regularize the recovery of superquadrics by minimizing the distance of the points lying on the superquadric surface to the object mesh. The color indicates the distance of each point from the mesh.}
      \label{sq_samples_fitting}
\end{figure}

\begin{figure}[t]
      \centering
      \begin{subfigure}[b]{0.31\columnwidth}
         \centering
         \includegraphics[width=\textwidth]{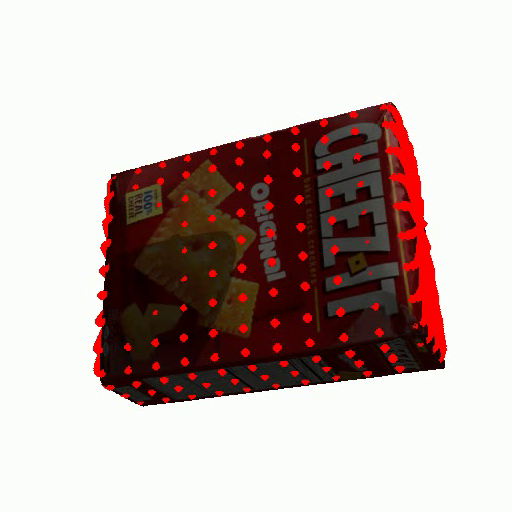}
     \end{subfigure}
     \hfill
     \begin{subfigure}[b]{0.31\columnwidth}
         \centering
         \includegraphics[width=\textwidth]{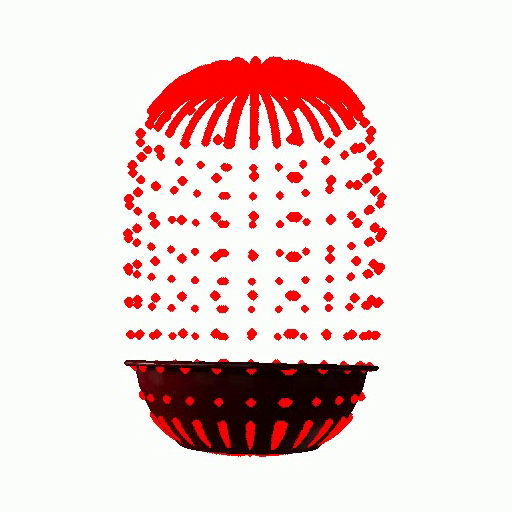}
     \end{subfigure}
     \hfill
     \vspace{0.3cm}
     \begin{subfigure}[b]{0.31\columnwidth}
         \centering
         \includegraphics[width=\textwidth]{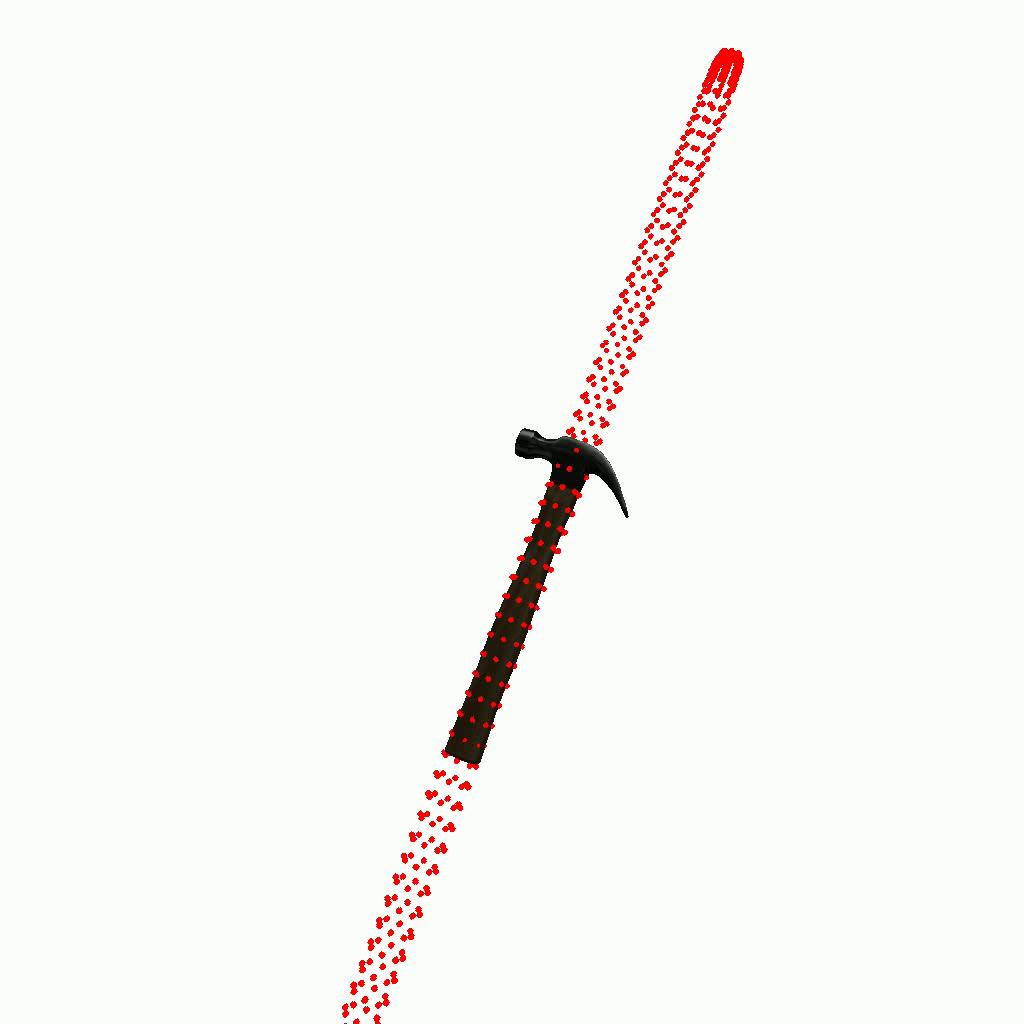}
     \end{subfigure}
     
     \begin{subfigure}[b]{0.31\columnwidth}
         \centering
         \includegraphics[width=\textwidth]{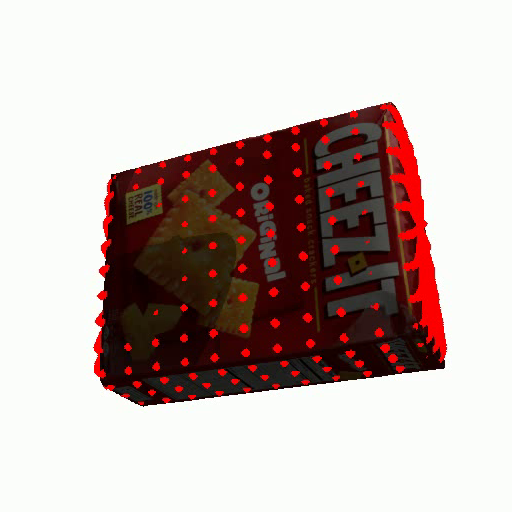}
     \end{subfigure}
     \hfill
     \begin{subfigure}[b]{0.31\columnwidth}
         \centering
         \includegraphics[width=\textwidth]{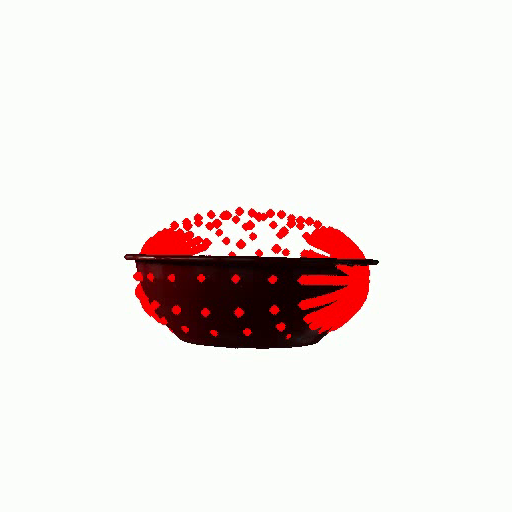}
     \end{subfigure}
     \hfill
     \begin{subfigure}[b]{0.31\columnwidth}
         \centering
         \includegraphics[width=\textwidth]{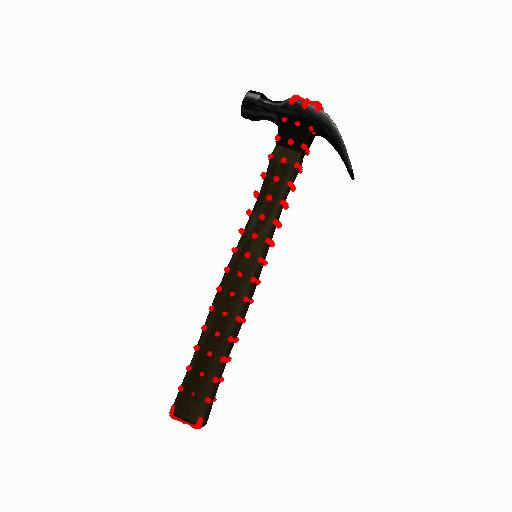}
     \end{subfigure}
      
      \caption{Superquadrics recovered for YCB objects: Top: Original EMS algorithm; Bottom: Our regularized version.}
      \label{sq_fitting}
\end{figure}

Accurate recovery of superquadrics is a difficult problem. Liu et al.~\cite{Liu2022} have recently proposed a robust algorithm for fitting superquadrics to point clouds. They formulate the recovery of superquadrics as a maximum likelihood estimation problem and propose a variation of the expectation-maximization algorithm that exploits the geometric features of superquadrics to solve it. Their approach is called Expectation, Maximization, and Switching (EMS). While excellent results can be obtained for objects whose shape is approximately within the geometric scope of superquadrics, objects outside this domain can produce misleading solutions. This can be seen in the top row of Figure \ref{sq_fitting}. While the cuboid object is recovered very well, the other objects lead to unsatisfactory results. This is because the solutions are found solely by optimizing the probability $p(\mathbf{x}_i; \bm{\theta})$ of points on the mesh surface $\mathbf{x}_i$ under the superquadric parameters $\bm{\theta}$. As can be seen for the second object, although a large region of the superquadric is far from the bowl, all points on the mesh surface are very close to the surface of the superquadric, which defines optimality in this case. While the authors propose to use an additional term in the loss function expressing a preference for superquadrics with smaller surface area, we have found that this negatively affects performance on objects that could otherwise be recovered accurately, as it biases the solution towards spherical parameterizations. Instead, we introduce a second loss term that minimizes the distance of points on the superquadric surface from the original mesh surface. The regularization effect can be seen in Figure \ref{sq_samples_fitting}. Figure \ref{sq_fitting} highlights the difference our adaption makes for recovering superquadrics on the YCB objects. It can be seen that the representation of the cuboid object remains accurate, while the approximations of objects that cannot be fitted precisely converge to more representative superquadric parameterizations.

\begin{table*}
\caption{Variations in lifting success rates for different representations of object geometry. For point clouds, the number after the dash indicates the number of points sampled on the object surface.}
\begin{center}
\begin{tabular}{c c c c c c c c c c} 
 \toprule
  & \multirow{2}{*}{Object} & \multicolumn{7}{c}{Success rate} \\  
  \cmidrule{3-9}
  & & COM & SDF & BBox & SQ & PC-32 & PC-128 & PC-512 \\
 \midrule

\multirow{2}{*}{\includegraphics[width=0.07\columnwidth, height=0.07\columnwidth]{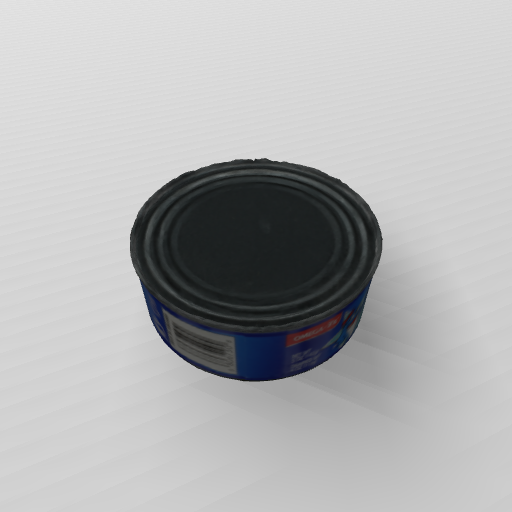} }& \multirow{2}{*}{007\_tuna\_fish\_can} & \multirow{2}{*}{$0.99 \pm 0.0$} & \multirow{2}{*}{$\bm{1.0 \pm 0.0}$} & \multirow{2}{*}{$\bm{1.0 \pm 0.0}$} & \multirow{2}{*}{$\bm{1.0 \pm 0.0}$} & \multirow{2}{*}{$\bm{1.0 \pm 0.0}$} & \multirow{2}{*}{$\bm{1.0 \pm 0.0}$} & \multirow{2}{*}{$\bm{1.0 \pm 0.0}$} \\ 
 & & & & & & & \\ 
\multirow{2}{*}{\includegraphics[width=0.07\columnwidth, height=0.07\columnwidth]{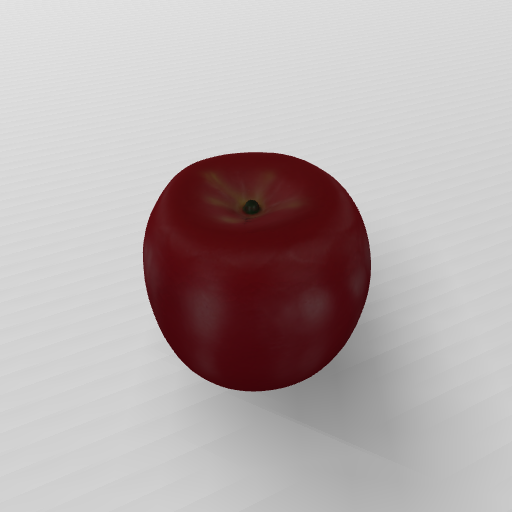} }& \multirow{2}{*}{013\_apple} & \multirow{2}{*}{$\bm{1.0 \pm 0.0}$} & \multirow{2}{*}{$\bm{1.0 \pm 0.0}$} & \multirow{2}{*}{$\bm{1.0 \pm 0.0}$} & \multirow{2}{*}{$\bm{1.0 \pm 0.0}$} & \multirow{2}{*}{$\bm{1.0 \pm 0.0}$} & \multirow{2}{*}{$\bm{1.0 \pm 0.0}$} & \multirow{2}{*}{$\bm{1.0 \pm 0.0}$} \\ 
 & & & & & & & \\  
\multirow{2}{*}{\includegraphics[width=0.07\columnwidth, height=0.07\columnwidth]{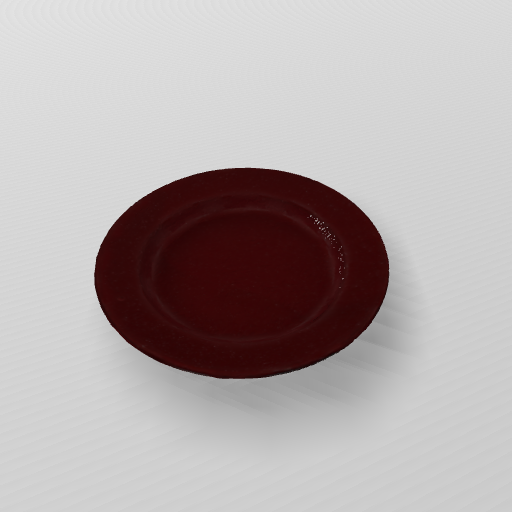} }& \multirow{2}{*}{029\_plate} & \multirow{2}{*}{$0.74 \pm 0.14$} & \multirow{2}{*}{$\bm{0.95 \pm 0.0}$} & \multirow{2}{*}{$0.93 \pm 0.02$} & \multirow{2}{*}{$\bm{0.95 \pm 0.04}$} & \multirow{2}{*}{$0.5 \pm 0.39$} & \multirow{2}{*}{$0.94 \pm 0.03$} & \multirow{2}{*}{$0.94 \pm 0.04$} \\ 
 & & & & & & & \\ 
\multirow{2}{*}{\includegraphics[width=0.07\columnwidth, height=0.07\columnwidth]{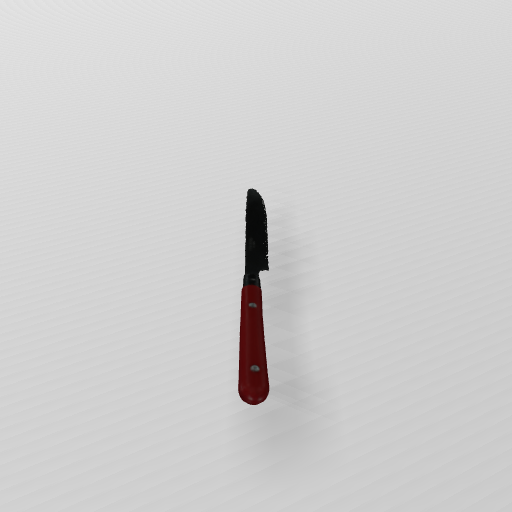} }& \multirow{2}{*}{032\_knife} & \multirow{2}{*}{$0.77 \pm 0.07$} & \multirow{2}{*}{$0.81 \pm 0.07$} & \multirow{2}{*}{$\bm{0.96 \pm 0.02}$} & \multirow{2}{*}{$0.85 \pm 0.1$} & \multirow{2}{*}{$0.71 \pm 0.02$} & \multirow{2}{*}{$0.9 \pm 0.02$} & \multirow{2}{*}{$0.79 \pm 0.06$} \\ 
 & & & & & & & \\ 
\multirow{2}{*}{\includegraphics[width=0.07\columnwidth, height=0.07\columnwidth]{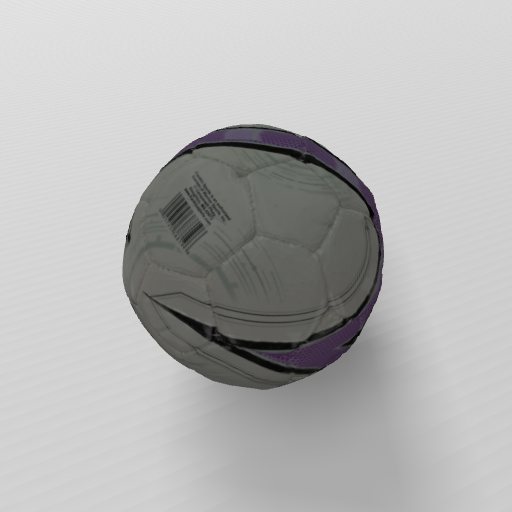} }& \multirow{2}{*}{053\_mini\_soccer\_ball} & \multirow{2}{*}{$\bm{1.0 \pm 0.0}$} & \multirow{2}{*}{$\bm{1.0 \pm 0.0}$} & \multirow{2}{*}{$\bm{1.0 \pm 0.0}$} & \multirow{2}{*}{$\bm{1.0 \pm 0.0}$} & \multirow{2}{*}{$\bm{1.0 \pm 0.0}$} & \multirow{2}{*}{$\bm{1.0 \pm 0.0}$} & \multirow{2}{*}{$\bm{1.0 \pm 0.0}$} \\ 
 & & & & & & & \\ 
\multirow{2}{*}{\includegraphics[width=0.07\columnwidth, height=0.07\columnwidth]{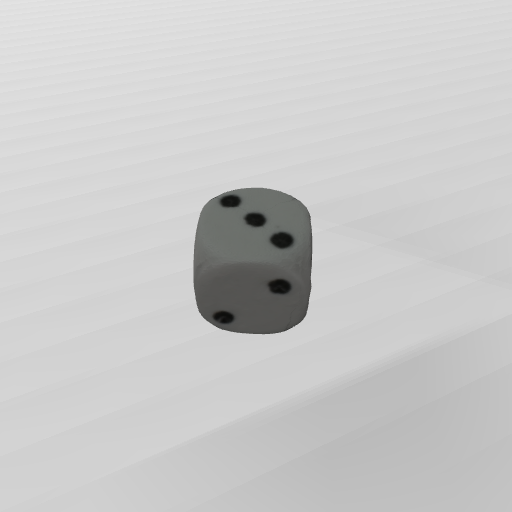} }& \multirow{2}{*}{062\_dice} & \multirow{2}{*}{$0.65 \pm 0.3$} & \multirow{2}{*}{$0.85 \pm 0.08$} & \multirow{2}{*}{$\bm{0.95 \pm 0.02}$} & \multirow{2}{*}{$0.94 \pm 0.01$} & \multirow{2}{*}{$0.55 \pm 0.07$} & \multirow{2}{*}{$0.86 \pm 0.0$} & \multirow{2}{*}{$0.6 \pm 0.12$} \\ 
 & & & & & & & \\ 
\multirow{2}{*}{\includegraphics[width=0.07\columnwidth, height=0.07\columnwidth]{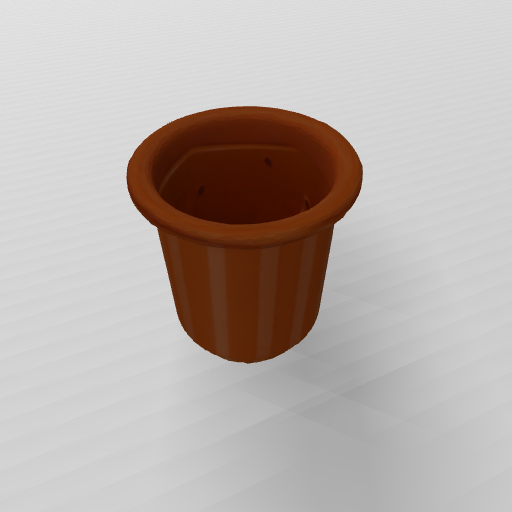} }& \multirow{2}{*}{065-a\_cups} & \multirow{2}{*}{$0.99 \pm 0.0$} & \multirow{2}{*}{$\bm{1.0 \pm 0.0}$} & \multirow{2}{*}{$\bm{1.0 \pm 0.0}$} & \multirow{2}{*}{$\bm{1.0 \pm 0.0}$} & \multirow{2}{*}{$\bm{1.0 \pm 0.0}$} & \multirow{2}{*}{$\bm{1.0 \pm 0.0}$} & \multirow{2}{*}{$\bm{1.0 \pm 0.0}$} \\ 
 & & & & & & & \\ 
\multirow{2}{*}{\includegraphics[width=0.07\columnwidth, height=0.07\columnwidth]{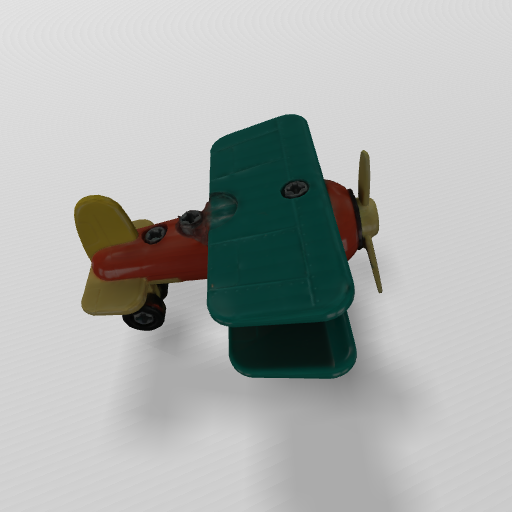} }& \multirow{2}{*}{072-a\_toy\_airplane} & \multirow{2}{*}{$0.93 \pm 0.02$} & \multirow{2}{*}{$\bm{0.99 \pm 0.01}$} & \multirow{2}{*}{$\bm{0.99 \pm 0.01}$} & \multirow{2}{*}{$0.97 \pm 0.03$} & \multirow{2}{*}{$0.95 \pm 0.01$} & \multirow{2}{*}{$\bm{0.99 \pm 0.0}$} & \multirow{2}{*}{$0.96 \pm 0.02$} \\ 
 & & & & & & & \\ 
\multirow{2}{*}{\includegraphics[width=0.07\columnwidth, height=0.07\columnwidth]{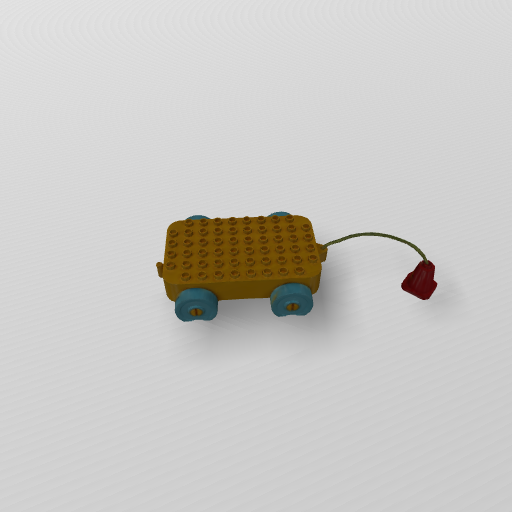} }& \multirow{2}{*}{073-g\_lego\_duplo} & \multirow{2}{*}{$0.9 \pm 0.01$} & \multirow{2}{*}{$0.97 \pm 0.01$} & \multirow{2}{*}{$\bm{0.99 \pm 0.0}$} & \multirow{2}{*}{$0.96 \pm 0.02$} & \multirow{2}{*}{$0.95 \pm 0.01$} & \multirow{2}{*}{$\bm{0.99 \pm 0.0}$} & \multirow{2}{*}{$0.98 \pm 0.01$} \\ 
 & & & & & & & \\ 
\multirow{2}{*}{\includegraphics[width=0.07\columnwidth, height=0.07\columnwidth]{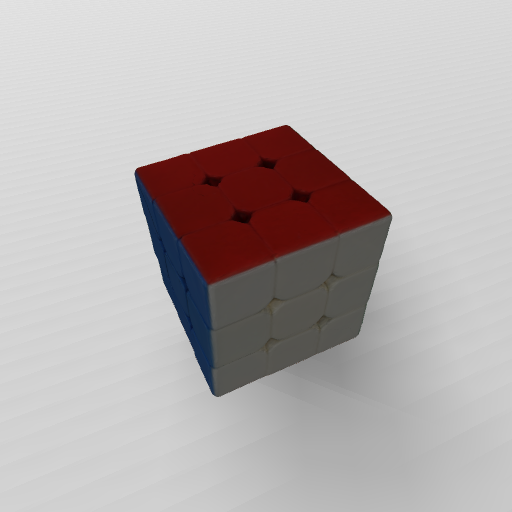} }& \multirow{2}{*}{077\_rubiks\_cube} & \multirow{2}{*}{$\bm{1.0 \pm 0.0}$} & \multirow{2}{*}{$\bm{1.0 \pm 0.0}$} & \multirow{2}{*}{$\bm{1.0 \pm 0.0}$} & \multirow{2}{*}{$\bm{1.0 \pm 0.0}$} & \multirow{2}{*}{$\bm{1.0 \pm 0.0}$} & \multirow{2}{*}{$\bm{1.0 \pm 0.0}$} & \multirow{2}{*}{$\bm{1.0 \pm 0.0}$} \\ 
 & & & & & & & \\ 

\midrule

\multicolumn{2}{c}{All 78 YCB objects} & $0.95 \pm 0.02$ & $0.96 \pm 0.01$ & $\bm{0.97 \pm 0.01}$ & $0.96 \pm 0.02$ & $0.92 \pm 0.03$ & $\bm{0.97 \pm 0.0}$ & $0.95 \pm 0.02$ \\

 \bottomrule
\end{tabular}
\end{center}
\label{object_specific_success_rates}
\end{table*}

\subsection{Implicit SDF Representation}
In addition to the explicit representations mentioned above, complex geometries can also be represented implicitly.
A signed distance function is a continuous mapping $\phi (\bm{x}): \mathbb{R}^3 \rightarrow \mathbb{R}$ that projects a 3D point $\bm{x}$ to its distance to a surface, where the sign indicates whether the point is inside ($-$) or outside ($+$) the surface.
To make this information available to a RL policy, we evaluate the SDF on a small set of points. Specifically, we query the distance to the object surface at the fingertip positions of the robot hand, resulting in a 5-dimensional observation $\irow{d(\bm{x}^{ft}_{1}, \delta \mathcal{O}),&\dots,&d(\bm{x}^{ft}_{5}, \delta \mathcal{O})}$, where $\bm{x}^{ft}_{i}$ is the position of the \textit{i}-th fingertip and $\delta \mathcal{O}$ is the object surface. Hence, the agent can perceive the distance of its fingers to the object surface and is thus implicitly informed about the geometry it is interacting with. 

Computing the signed distance of a set of points to a high-fidelity mesh is expensive~\cite{Narang2022}, however, and would be required in every environment at every step of the simulation. To nevertheless make this observation type feasible for massively parallel reinforcement learning, we precompute a discrete voxel-based representation of the SDF for all object meshes (see Figure~\ref{sdf}). Specifically, we create a $200 \times 200 \times 200$ grid filling the space $[-0.5 m, 0.5 m]^3$ around the object center. This creates a spatial resolution of 5\,mm, which we found to be sufficient in our experiments. When query points lie outside the volume of the voxel-grid, we return the result from the closest voxel. With this approach, the lookup of signed distances to the robot fingertips takes only 0.31\,ms for all 16,384 parallel environments, leading to negligible impact on the overall training performance. SDF voxel-grids computed once for an object mesh are stored permanently to enable fast reloading in future training runs.

\subsection{SDF-based Reward Shaping}

Using our SDF-based observation of the distance between end-effector and object, we introduce a reward term that incentives the agent to make maximum contact with whatever object it interacts with. Specifically, we reward minimizing the distance of the fingertips to the object surface:
\begin{equation}
    r_{sdf} = \frac{1}{d_t + \epsilon_{sdf}},
\end{equation}

where $d_t = \sum_{i} d(\bm{x}^{ft}_i, \delta \mathcal{O})$ and $\epsilon_{sdf}$ is a constant. We hypothesize that this will help reducing the search space for the policy, while being unbiased and object-independent, since no particular position or orientation is imposed for grasping an object, only that the agent should make contact.

\subsection{Policy Optimization}

We use proximal policy optimization (PPO)~\cite{Schulman2017} to learn the agents, since the algorithm has been show to converge robustly on problems with high-dimensional action spaces, and make use of the high-performance implementation provided by Makoviichuk and Makoviychuk~\cite{rl-games2022}. We run 16,384 parallel environments to collect data. After every 32 steps, we update the policy for 5 epochs on the rollout data with a batch size of 32,768. All experiments are run on a single NVIDIA RTX A6000 GPU with 48GB of memory. The policy is computed by a three-layer fully connected neural network, as detailed in Figure~\ref{policy_architecture}.

\section{Experimental Setup}

We use NVIDIA Isaac Gym~\cite{Makoviychuk2021} as our physics simulator. The robot consists of a UR5e arm and the Schunk SIH hand, which features 11 joints, the actuation of which is coupled, leading to five controllable degrees of freedom (DoF). 

\subsection{Observations, Actions, and Rewards}
Observations consist of the current pose of the robot hand and its joint positions as well as a representation of the object. For targeted picking from cluttered bins, $\bm{o}_t$ represents the observation of the target object only. Hence, we do not inform the policy about other objects or the walls of the bin. In this scenario, a random object in the bin is selected as the target at the beginning of each episode. 

The agent controls the robot via 11-dimensional actions, representing the desired change in position and rotation of the hand pose and the controllable DoF positions of the hand. From the desired pose of the robot hand, we compute the UR5e joint position targets using the Jacobian transpose method. The action frequency is 30\,Hz. 

An episode terminates when the object falls off the table, is lifted to the target height, or the policy runs out of time after 300 actions taken. We use the following reward function to train the lifting policies:
\begin{equation}
    r(s_t, a_t) = c_1 \frac{1}{|\overline{h} - \Delta h_t| + \epsilon_h} + c_2 \mathds{1}(\Delta h_t \ge \overline{h}) + c_3 r_{sdf}
    \label{reward_function}
\end{equation}
where $\Delta h_t$ is the height an object has been lifted by, relative to its starting position, and $\overline{h}$ is the target value for $\Delta h_t$. The first two reward terms incentivize the policy to lift the object up, while the last term reduces the search space by encouraging the agent to make contact with the object. We set $\overline{h}$ to 0.2\,m for our experiments and weigh the terms with $c_1 = 0.5$, $c_2=5000$, and $c_3=1$, where $\epsilon_h = 0.02$ and $\epsilon_{sdf} = 0.025$.
We evaluate the success of a policy rollout by whether an object was raised by $\overline{h}$, whereupon the episode terminates. We report the mean and standard deviation of three seeds for all experiments.

\subsection{Object Dataset}
We consider the problem of grasping and lifting objects of variable shape and size. To this end, we evaluate our policies on all 78 YCB object models~\cite{Calli2015} for which laser scans exist. The weights of all objects are set to realistic values and their inertia tensors are computed from the geometry in Isaac Gym. The objects are spawned in random poses and then dropped onto a table to create realistic initial configurations. 

\begin{table}[t]
\caption{Sample efficiency of different object representations measured by the number of environment steps (in million) and the runtime in minutes required to reach success rate thresholds.}
\begin{center}

\resizebox{\columnwidth}{!}{
\begin{tabular}{c c c c c} 
 \toprule
  \multirow{2}{*}{\makecell{Object \\repr.}} & \multicolumn{2}{c}{90\,\%} & \multicolumn{2}{c}{95\,\%} \\
  \cmidrule{2-3}
  \cmidrule{4-5}
  & Steps & Runtime [min] & Steps & Runtime [min] \\
 \midrule
 COM & $149 \pm 30$ & $47 \pm 9$ & $550 \pm 161$ & $116 \pm 34$ \\
 SDF & $\bm{134 \pm 5}$ & $72 \pm 3$ & $276 \pm 51$ & $152 \pm 28$ \\
 BBox & $165 \pm 8$ & $69 \pm 3$ & $274 \pm 44$ & $127 \pm 23$ \\
 SQ & $178 \pm 18$ & $58 \pm 6$ & $436 \pm 52$ & $220 \pm 26$ \\
 PC-32 & $196 \pm 15$ & $52 \pm 4$ & $474 \pm 170$ & $178 \pm 63$ \\
 PC-128 & $173 \pm 4$ & $\bm{47 \pm 2}$ & $\bm{232 \pm 1}$ & $\bm{65 \pm 1}$ \\
 PC-512 & $219 \pm  19$ & $62 \pm 6$ & $277 \pm 12$ & $87 \pm 4$ \\
 
 \bottomrule
\end{tabular}
}
\end{center}
\label{sample_efficiency_of_object_representations}
\end{table}

\section{Results}

We evaluate the impact of geometry-awareness and the type of object representation on the performance of RL-based grasping policies. Further, we analyze the effect of our geometry-aware shaped reward on the learning behavior. Finally, because we observe that PPO is able to learn robust grasping policies that naturally exhibit useful behaviors such as pre-grasp manipulation and re-grasping, we test whether our method can be straightforwardly applied to picking objects from cluttered bins.

\subsection{Role of Object Representation}

In the following, we discuss the role of object representations in learning grasping policies. We compare the performance that the different representations achieve during training both in terms of sample efficiency and final performance to capture the strengths and weaknesses of both succinct and expressive representations. 

Table~\ref{object_specific_success_rates} reports the final test success rate, both overall and for specific objects, after training for a total of 500 epochs, where each epoch consists of 32 steps in 16,384 parallel environments. It can be seen that the geometry-independent policies provide a very strong baseline, even for grasping objects with significantly different shapes and sizes. Nevertheless, we were able to demonstrate a performance advantage from geometry awareness. 
Especially for very small objects, such as the dice, or objects far from cubic shape, such as the plate or knife, geometry-awareness  enables learning of specific grasping strategies.
While observations of the distance of the fingertips from the object surface learn more quickly, their final performance falls short of that reached via bounding boxes or point clouds. Bounding box representations showed strong final performance while maintaining high convergence speed. They introduce the geometric information necessary to learn a proficient object-specific policy for the studied task. Superquadrics reach a final performance just short of the bounding box policies. It should be noted that superquadrics subsume the cuboid representation of objects via bounding boxes. While they are in principle strictly more expressive, using superquadrics introduces the added challenge of fitting them to a given object mesh. While they are able to fit many common objects precisely, superquadric representations may be misleading for objects that cannot be approximated well. Despite our efforts to ensure that meaningful superquadric parameterizations were recovered for all objects, the resulting policies performed worse overall than the bounding box policies. For point cloud-based observations, we encountered a strong dependence of performance on the number of sampled points. For only 32 points, this observation type did not benefit the policy and actually leads to a reduction in final performance, compared to the geometry-independent baseline. As more points are sampled, training performance  improves, but speed of convergence goes down. We found the best trade-off to be at 128 points sampled per object, resulting in a policy that converges quickly and to the shared highest overall performance.

\begin{figure}[t]
      \centering
      \includegraphics[width=\columnwidth]{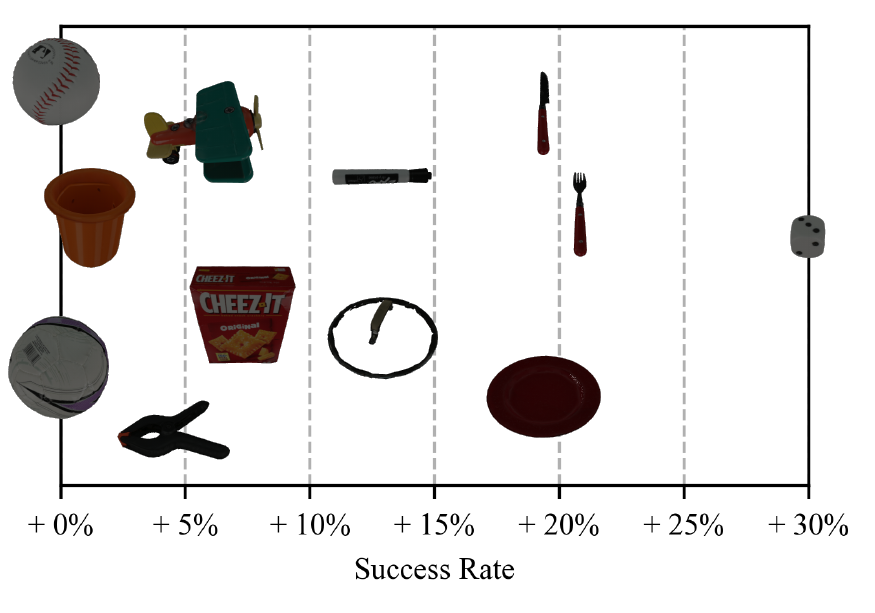}
			\caption{Performance gains of geometry-aware (BBox) policy over the COM baseline. While there are little to no improvements for cubic or spherical objects of average extent, significant performance improvements can be achieved for objects of irregular shape and size. Visualization inspired by Huang et al.~\cite{Huang2021}.}
      \label{object_performance_gain}
\end{figure}

Table~\ref{sample_efficiency_of_object_representations} reports the number of samples required for the policies to achieve an average success rate of 90\,\% or 95\,\% across all objects during training. It can be seen, that the proposed implicit representation of object geometry via signed distances reaches the success rate threshold of 90\,\% first. The representation of object via 128 sampled points already outperforms all others when it comes to reaching a success rate of 95\,\%. Learning from point clouds with only 32 samples does not adequately capture object geometry, resulting in unstable training performance and thus significantly worse performance compared to the other point cloud runs. While the geometry-agnostic representation by the object's center of mass is able to reach the first threshold quickly, improving performance above 90\,\% becomes challenging, given the limited information available. 


While the gains achieved by incorporating object shape may seem minor in terms of the success rate average across the entire object set, it is important to note that the COM baseline works very well for the majority of objects studied. The most interesting comparisons can be made for objects of irregular shape and size. We find that geometry-awareness provides significant improvements for these object types where a universal, object-independent strategy fails. Figure~\ref{object_performance_gain} shows this performance gain for specific objects. It highlights the correlation of performance gained through shape-awareness and object irregularity. The dice, for which the biggest improvement can be made is the smallest object in the entire dataset. The fork and knife, which also profit significantly from shape-awareness, have some of the largest aspect ratio of their extent along the three principal axes. 

\begin{figure}[t]
      \centering
      \includegraphics[width=\columnwidth]{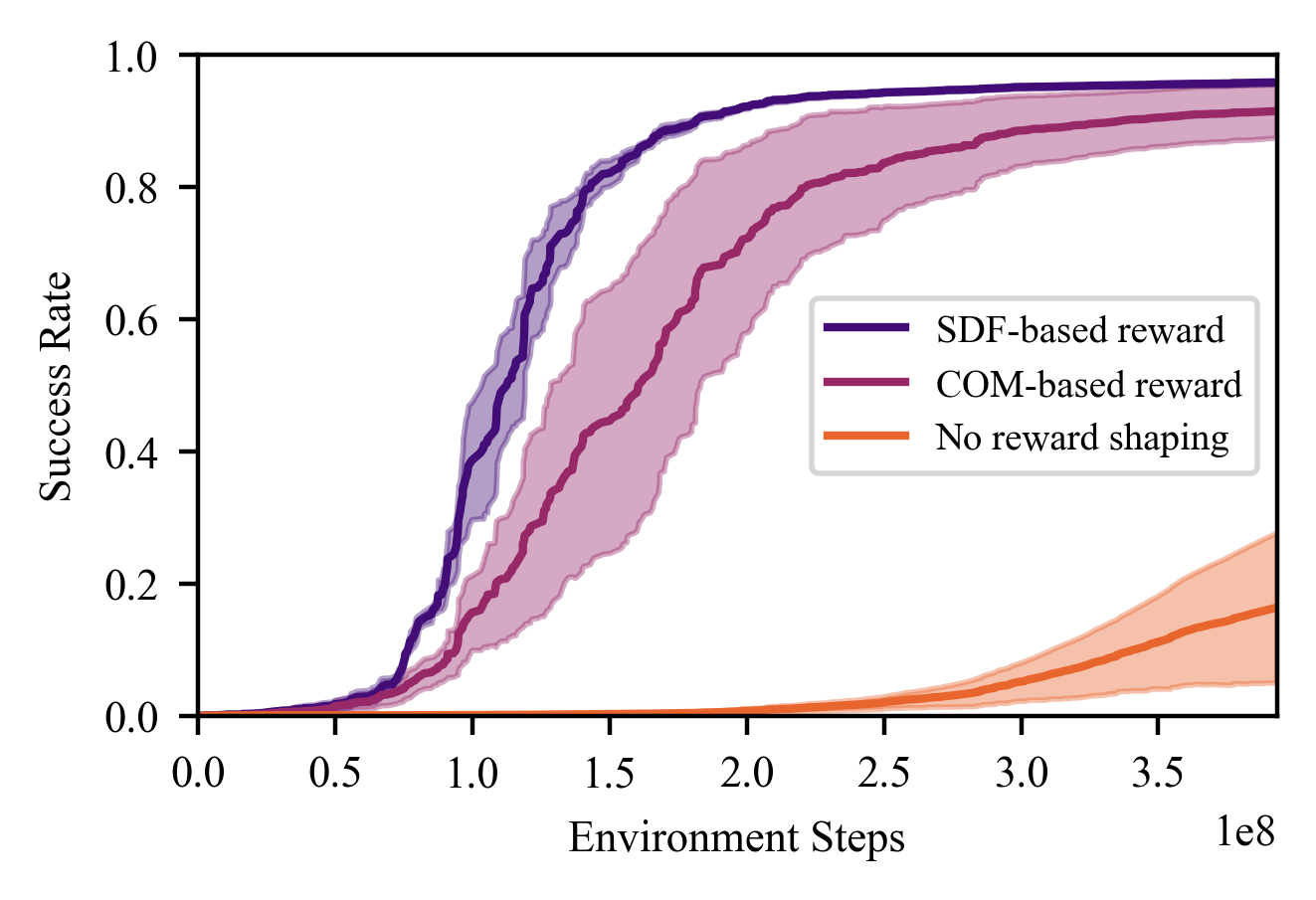}
      \caption{Average success rate achieved during training with different reward functions. }
      \label{sdf_reward_training_performance}
\end{figure}

\subsection{SDF-based Reward Shaping}

Next, we benchmark our proposed SDF-based reward component against two ablations. The comparative training performance is shown in Figure~\ref{sdf_reward_training_performance}. 

Not rewarding the policy based on the position of its end-effector relative to the object, i.e. setting $c_3 = 0$ in Eq.~\ref{reward_function}, leads to a drastic deterioration in performance, as most transitions do not receive a meaningful reward signal, which hinders learning progress. 
In a second ablation, we replace the distance of the fingers to the object surface by their distance to its center of mass. This represents a naive formulation of our contact-inducing reward. The results show that this leads to accelerated learning, since the agent is encouraged to move its hand close to the object. However, this simplification introduces bias in how to grasp an object, motivating the policy to always reach for the center of mass. Especially for larger or more irregularly shaped objects, this strategy becomes infeasible. Consequently, the COM-based reward shaping performs inferior to our proposed SDF-based reward both in terms of sample efficiency and final performance. Thus, we can confirm that the novel SDF-based reward induces the desired effect of guiding the search effectively, while being agnostic to how an object is grasped by the policy. This leads to faster convergence and improved final performance when grasping the YCB objects.

\subsection{Application to Cluttered Bin-Picking}

\begin{figure}[t]
      \centering
      
      \begin{subfigure}[b]{0.325\columnwidth}
         \centering
         \includegraphics[width=\textwidth]{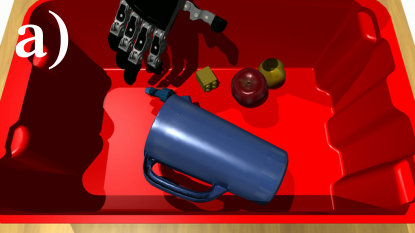}
     \end{subfigure}
     \hfill
     \begin{subfigure}[b]{0.325\columnwidth}
         \centering
         \includegraphics[width=\textwidth]{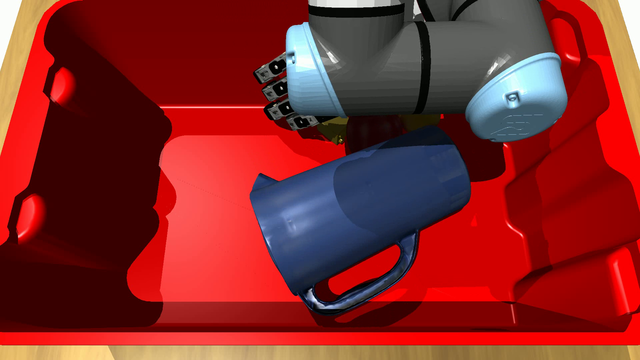}
     \end{subfigure}
     \hfill
     \begin{subfigure}[b]{0.325\columnwidth}
         \centering
         \includegraphics[width=\textwidth]{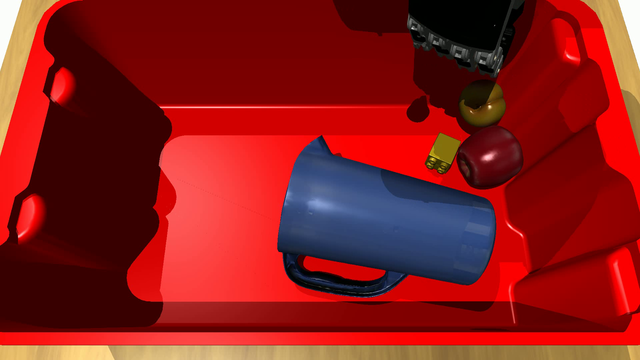}
     \end{subfigure}
     
     \vspace{0.25cm}
     
     \begin{subfigure}[b]{0.325\columnwidth}
         \centering
         \includegraphics[width=\textwidth]{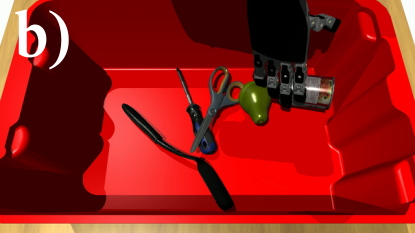}
     \end{subfigure}
     \hfill
     \begin{subfigure}[b]{0.325\columnwidth}
         \centering
         \includegraphics[width=\textwidth]{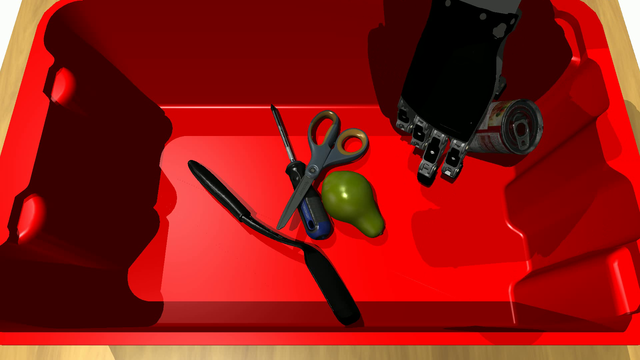}
     \end{subfigure}
     \hfill
     \begin{subfigure}[b]{0.325\columnwidth}
         \centering
         \includegraphics[width=\textwidth]{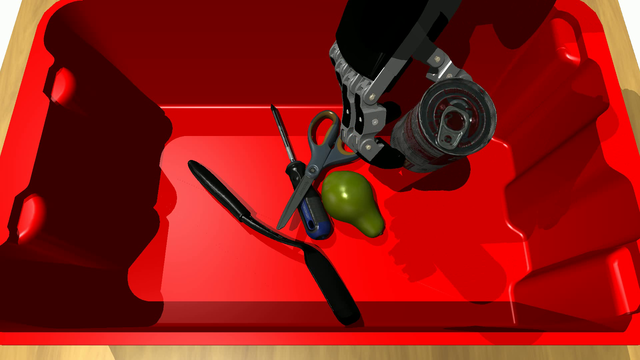}
     \end{subfigure}

     \vspace{0.1cm}

      \caption{Strategies learned for targeted bin picking: (a) The agent lifts the tiny airplane part by sliding it up the near wall; (b) The tomato soup can is initially in a position that is difficult to reach. Through continuous manipulation, the policy succeeds in bringing it into an upright pose from which it can be grasped.}
      \label{bin_picking_behaviors}
\end{figure}

To analyze whether the interactive policies can utilize pre-grasp manipulation and regrasping to solve more demanding tasks, we apply our method to targeted picking from cluttered bins. To initialize a scenario, we drop five randomly chosen YCB objects into a bin. The agent is tasked to retrieve a specific object from this cluttered bin. We provide the policy with the same information as for the unobstructed grasping of objects lying on the table. Hence, only proprioceptive measurements of the robot's state and the state of the target object are included in the observation. We represent the target object via its bounding box, since stong performance was obtained for this low-dimensional observation type in the single object case. The trained policy achieves a success rate of $0.85 \pm 0.02$, averaged over all objects. When we increase task difficulty by dropping ten random objects in the bin, the same policy still achieves a success rate of $0.81 \pm 0.04$, underlining the robustness of the strategies that have been learned.
Interesting behaviors that emerged in this setting are shown Figure~\ref{bin_picking_behaviors}. They include extended pre-grasp interactions to reorient objects and the utilization of environmental constraints, such as the bin walls.

Naturally, the policy cannot reach quite the same performance as for the single object case. Nevertheless, the results show that our method is able to learn a proficient controller even for this challenging setting. This approach can serve as a strong baseline for future work on targeted picking from cluttered bins that explicitly takes the context into account.

\section{Conclusions and Discussion}

Our work showed that regular on-policy optimization can train a single agent that is able to grasp various geometrically diverse objects. We confirm that learning with shape-independent observations can bring about a universal strategy that achieves strong performance. Nevertheless, simple geometry-conditioned observations can boost the performance significantly for objects of irregular shape and size, leading to more generally capable policies. Basic observations such as bounding boxes proved sufficient for this task and offered a strong trade-off between expressiveness and compactness in our experiments.
Point clouds showed very strong training performance when the right number of surface points is sampled. While a higher number of sampled points requires increased training times, combining reinforcement learning with a pre-trained point-cloud encoder might alleviate this issue and presents an interesting avenue for future work. Since the learned agents exhibit robust interactive grasping strategies, we evaluated the performance of our framework for the unstructured scenario of retrieving a specific object from a cluttered bin. Even without any context information we were able to learn proficient interactive policies for this task that reorient objects and utilize environmental constraints. Our framework can therefore serve as a proficient baseline for future work that explicitly considers object interactions and environment representations. Robust inference of succinct object descriptions from data available in the real world, such as RGB-D images, remains an open problem and is left to future work.

Furthermore, we introduced a novel geometry-aware reward function that benefits both the speed of convergence as well as final performance of the interactive grasping policies. Minimizing the distance between the end-effector and object surface proved to be an essential addition, as it effectively guides the search without imposing limitations on the specific grasping strategy the policy finds. We demonstrated that signed distances required to evaluate this reward can be approximated efficiently, making it readily applicable for high-performance RL. Hence, this formulation of the reward can potentially be a valuable addition for reinforcement learning of grasping policies in general, for example, for different gripper morphologies or to promote stable grasps when using tools.

\section*{Acknowledgment}
\noindent \footnotesize{This work has been funded by the German Ministry of Education and Research (BMBF), grant no. 01IS21080, project “Learn2Grasp: Learning Human-like Interactive Grasping based on Visual and Haptic Feedback”.}

\addtolength{\textheight}{-11cm}   





\printbibliography

\end{document}